\documentclass{article}

\usepackage{microtype}
\usepackage{graphicx}
\usepackage[dvipsnames]{xcolor}
\usepackage{booktabs} 

\usepackage{hyperref}

\usepackage{pythonhighlight}

\usepackage[accepted]{icml2024}

\usepackage{amsmath}
\usepackage{amssymb}
\usepackage{mathtools}
\usepackage{amsthm}
\usepackage{multirow}
\usepackage{graphicx}

\usepackage{url}
\usepackage{booktabs}
\usepackage{color}
\usepackage{colortbl}
\usepackage{makecell}
\usepackage{graphicx}
\usepackage{soul}
\usepackage{float}
\usepackage{subfig}
\usepackage{overpic}
\newcommand{\modelname}{{\textsc{UltraFuser}}}
\newcommand{\dataname}{{\textsc{UltraChat 2}}}
\definecolor{mycolor1}{RGB}{249,243,243}
\definecolor{mycolor2}{RGB}{237,243,243}
\definecolor{mycolor3}{RGB}{238, 237, 245}
\definecolor{myblue}{rgb}{0.843,0.898,0.941}
\definecolor{mygrey}{rgb}{0.976,0.976,0.976}

\definecolor{mygreen}{HTML}{FCF0D9}
\definecolor{mybackblue}{HTML}{DDF0F5}

\definecolor{themegreen}{HTML}{C5E0B4}
\definecolor{themeblue}{HTML}{CFE2F3}
\definecolor{themeyellow}{HTML}{FFE699}
\definecolor{themedarkyellow}{HTML}{FFBF87}

\newcommand\figcaption{\def\@captype{figure}\caption}
\newcommand\tabcaption{\def\@captype{table}\caption}

\usepackage[capitalize,noabbrev]{cleveref}

\theoremstyle{plain}

\theoremstyle{definition}

\theoremstyle{remark}

\usepackage[textsize=tiny]{todonotes}

\icmltitlerunning{Mastering Text, Code and Math Simultaneously via  Fusing Highly Specialized Language Models}

\begin{document}

\twocolumn[
\icmltitle{Mastering Text, Code and Math Simultaneously via  Fusing \\  Highly Specialized Language Models}



\icmlsetsymbol{equal}{*}

\begin{icmlauthorlist}
\icmlauthor{Ning Ding}{equal,Tsinghua University}
\icmlauthor{Yulin Chen}{equal,Tsinghua University}
\icmlauthor{Ganqu Cui}{Tsinghua University}
\icmlauthor{Xingtai Lv}{Tsinghua University}
\icmlauthor{Weilin Zhao}{Tsinghua University}\\
\icmlauthor{Ruobing Xie}{Tsinghua University}
\icmlauthor{Bowen Zhou}{Tsinghua University}
\icmlauthor{Zhiyuan Liu}{Tsinghua University}
\icmlauthor{Maosong Sun}{Tsinghua University}

\end{icmlauthorlist}

\icmlaffiliation{Tsinghua University}{Tsinghua University}

\icmlcorrespondingauthor{Bowen Zhou}{}
\icmlcorrespondingauthor{Zhiyuan Liu}{liuzy}
\icmlcorrespondingauthor{Maosong Sun}{}

\icmlkeywords{Machine Learning, ICML}

\vskip 0.3in
]



\printAffiliationsAndNotice{\icmlEqualContribution} 

\begin{abstract}
Underlying data distributions of natural language, programming code, and mathematical symbols vary vastly, presenting a complex challenge for large language models (LLMs) that strive to achieve high performance across all three domains simultaneously.
Achieving a very high level of proficiency for an LLM within a specific domain often requires extensive training with relevant corpora, which is typically accompanied by a sacrifice in performance in other domains. In this paper, we propose to fuse models that are already highly-specialized directly. The proposed fusing framework, \modelname, consists of three distinct specialists that are already sufficiently trained on language, coding, and mathematics. A token-level gating mechanism is introduced to blend the specialists' outputs. A two-stage training strategy accompanied by balanced sampling is designed to ensure stability.
To effectively train the fused model, we further construct a high-quality supervised instruction tuning dataset, \dataname, which includes text, code, and mathematical content. This dataset comprises approximately 300,000 instructions and covers a wide range of topics in each domain. Experiments show that our model could simultaneously achieve mastery of the three crucial domains.

\end{abstract}

\section{Introduction}
\label{sec:intro}
If a piece of information can be serialized and tokenized, it is likely to be handled by large language models (LLMs)~\citep{bommasani2021opportunities,brown2020language,openai2023gpt}. 
LLMs, as one of the most advanced manifestations of artificial intelligence, have demonstrated proficiency in three representative symbol systems that are essential to human progress: natural language~\citep{ouyang2022training, bai2022constitutional}, which forms the cornerstone of human interaction; programming code~\citep{li2023starcoder,roziere2023code}, the backbone of our digital ecosystem; and mathematical reasoning, the framework underpinning scientific advancement~\citep{luo2023wizardmath,yang2023gpt}. 
The mastery of three domains would equip LLMs with unparalleled versatility. 
However, the intrinsic variability of data distribution across these domains presents a formidable challenge for an LLM to achieve consistently high performance \textit{at the same time}. One awkward situation is that it is challenging to integrate professional-level coding and mathematical abilities into a general conversational language model without loss. In other words, these skills are more often reflected in the numbers on related benchmarks rather than a real-world user interface.

\begin{figure*}[htbp]    
  \centering            
  \subfloat[Text-specialized model.]   
  {
      \label{fig:subfig1}\includegraphics[width=0.24\textwidth]{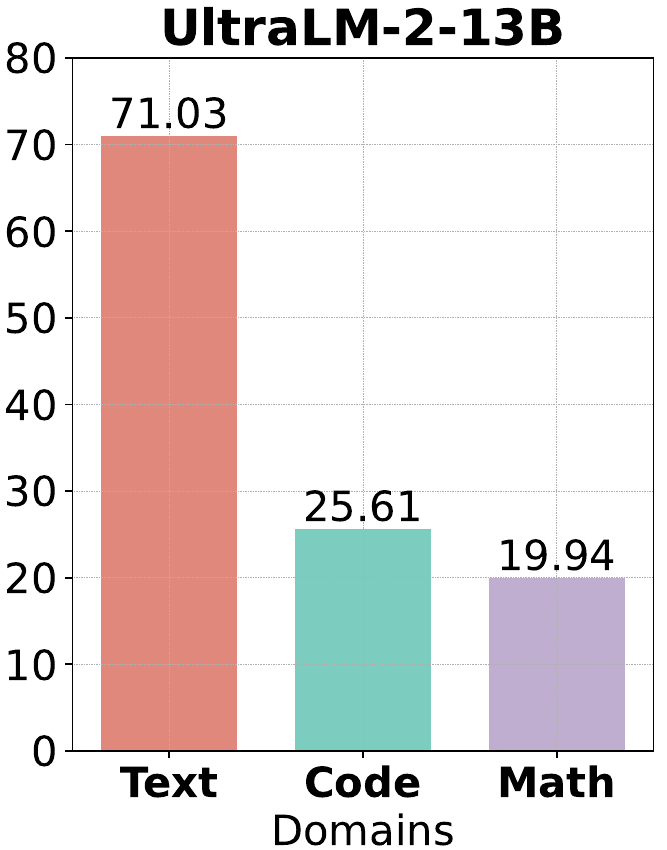}
  }
  \subfloat[Code-specialized model.]
  {
      \label{fig:subfig2}\includegraphics[width=0.24\textwidth]{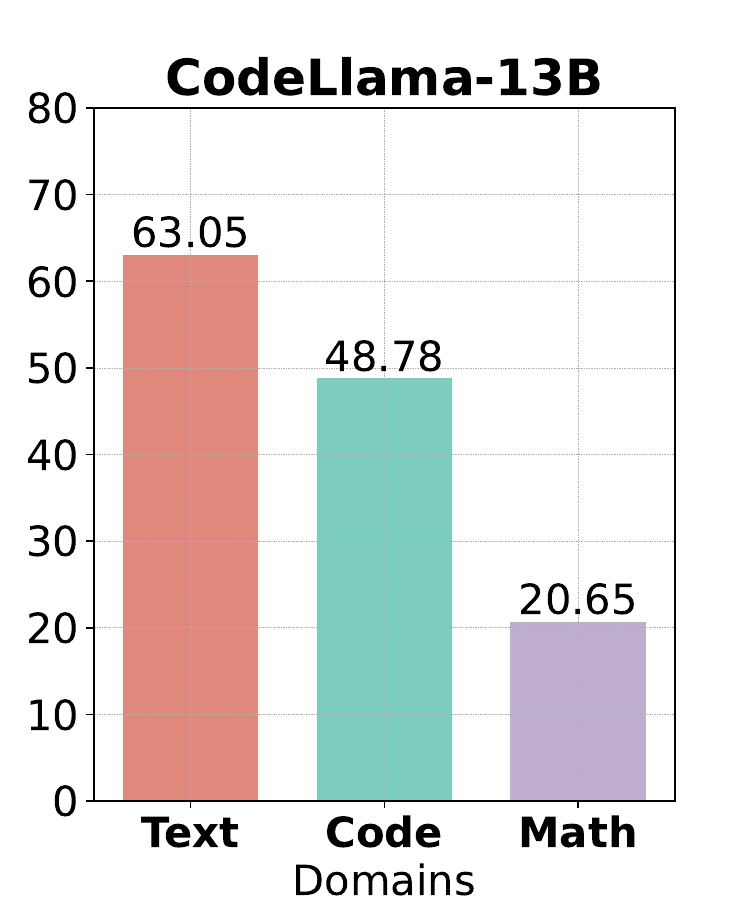}
  }
    \subfloat[Math-specialized model.]
  {
      \label{fig:subfig3}\includegraphics[width=0.24\textwidth]{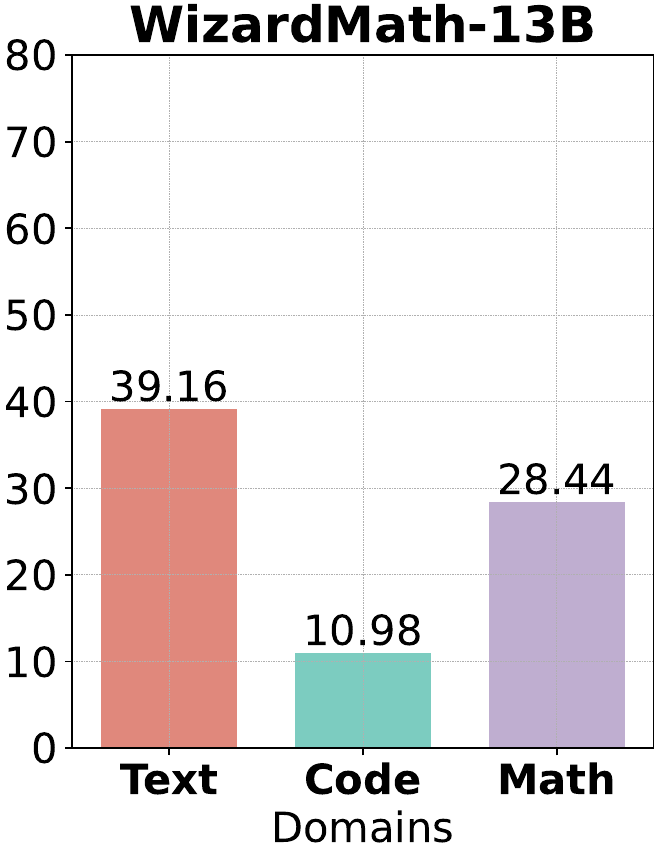}
  }
    \subfloat[Fused model.]
  {
      \label{fig:subfig4}\includegraphics[width=0.24\textwidth]{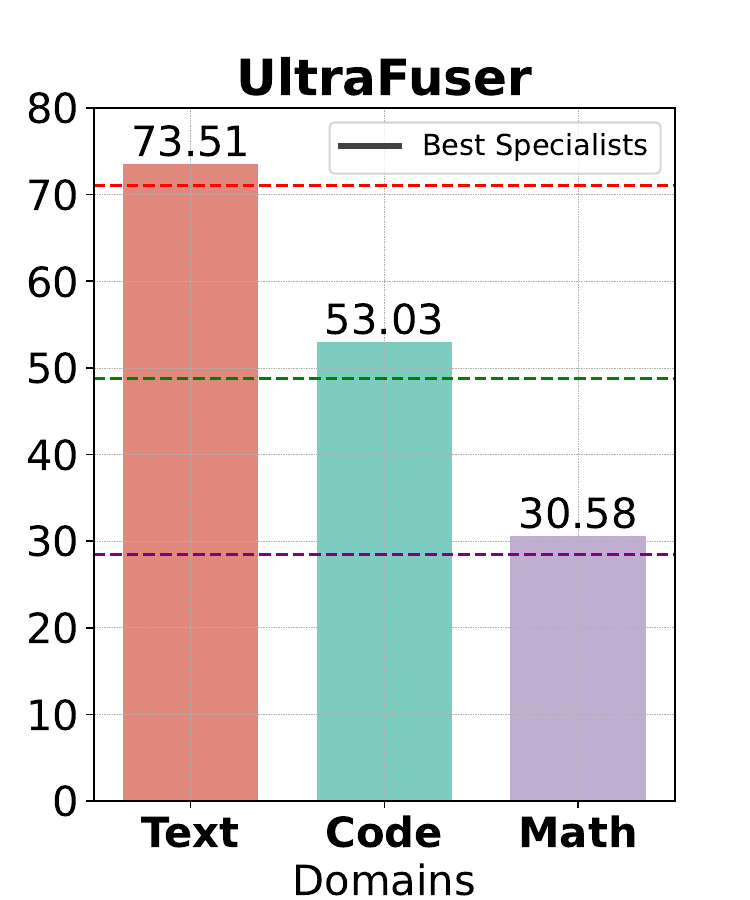}
  }
  \caption{Performance on three different domains of specialized models and our \modelname. 
  The performance for the text domain is computed by the average results on TruthfulQA (Acc)~\cite{lin2021truthfulqa} and AlpacaEval (Win Rate)~\cite{alpaca_eval} datasets; the performance for the code domain is Pass@1 of HumanEval~\cite{chen2021evaluating}; and the performance for the math domain is the average result of GSM8K (Pass@1)~\cite{cobbe2021training}, MATH (Pass@1)~\cite{hendrycks2021measuring}, SAT-Math (Acc) ~\cite{zhong2023agieval}, and AQuA-RAT (Acc)~\cite{ling-etal-2017-program} datasets. All numbers are zero-shot results.}
  \label{fig:intro}            
\end{figure*}

\looseness=-1 Figure~\ref{fig:intro} (a-c) demonstrates such a struggle by presenting the performance of three specialized models on the aforementioned domains, all initially based on the Llama-2~\citep{touvron2023llama} 13B architecture. Our findings reveal a clear trade-off: specialized training in one domain often comes at the expense of performance in the others, whereas training on all three types of data at the same time results in a simultaneous suboptimal situation. Delving into this situation, such an issue may be partially mitigated by careful designs of data engineering, training strategy, or prompt construction. 
However, in general, semantics in language, logic and structures in code, and abstract
symbol manipulations in math intricately always create a situation of mutual weakening. 
To elaborate further, comparing highly specialized models (such as those for coding or mathematics) with general-purpose models capable of performing all tasks (like GPT-4) for their expertise is a trap that can easily lead to misinformation.

This paper hopes to integrate specialized abilities into a general chat language model with as little loss as possible. More specifically, we propose to leverage separate models that are already highly specialized via a fusing structure. In this fusing framework, namely \modelname, we use three well-trained LLMs as initial specialist models in text, code, and math.~\footnote{Although we treat text, code, and math as three separate domains in this paper according to their symbol systems, 
they are not strictly segregated. For example, language can partially encompass the other two. This is discussed in Appendix~\ref{appendix:limitations}. } 
To ensure that the fused model benefits from the specialized knowledge of each specialist model, a dynamic gating mechanism is implemented, which sits on top of the three specialists and adaptively controls the contribution of each specialist to the final output logits based on the input data. Such a mechanism is adopted at the token level, which allows both the specialization of individual specialists and the generalization of the fused model.
The key to functioning the model is to train the gating module.
For example, when the model conducts code generation, we want the coding specialist to contribute more than the other two.
This necessitates a mixed instruction tuning dataset that contains the three domains for the training. Unlike language data, high-quality instruction-tuning datasets for code and math are scarcer in the open-source community. Inspired by \textsc{UltraChat}~\citep{ding2023enhancing}, we construct a comprehensive, diverse dataset with high quality, \dataname, to facilitate the development of advanced LLMs with the aforementioned expertise. \dataname \  contains 300,000 diverse and high-quality data (each part has 100,000), which are derived from 72 meta-topics and 1587 sub-topics.

Experiments show that highly specialized models may counter collapse if they are directly further trained, but we can effectively integrate their highly professional abilities into a general chat interface via \modelname. 
By training a fused model with UltraLM-2-13B, CodeLlama-13B, and WizardMath-13B as the specialists for three domains, we achieve consistently effective performance on seven benchmarks across language understanding, code generation, and mathematical reasoning. 
Our proposed model, data, training, and inference frameworks will be publicly available.

\section{Related Work}
\textbf{Large Language Models for Language.} 
With the proliferation of model parameters, enhancements in training data augmentation both in terms of quantity and quality, and continuous refinements in training algorithms, LLMs have exhibited an enhancement in language understanding, generation, and generalization capabilities.
These LLMs exhibit remarkable proficiency in accomplishing a wide array of natural language processing tasks, and showcase formidable capabilities in in-context learning and few-shot learning~\citep{brown2020language, ouyang2022training, openai2023gpt, chowdhery2022palm, zhang2022opt, touvron2023llama, taori2023alpaca, vicuna2023, xu2023wizardlm, ding2023enhancing, jiang2023mistral}. Despite originating from NLP tasks, as LLMs evolve, the boundaries between NLP tasks are gradually becoming blurred.

\looseness=-1 \textbf{Large Language Models beyond Language.} LLMs excel in processing various symbol systems including code, math symbols, DNA, and protein sequences. Models like StarCoder~\citep{li2023starcoder} and CodeLlama~\citep{roziere2023code}, trained on vast code repositories and interactions, are adept at code generation, bug fixing, and explanation~\citep{gpt-neo, gpt-j, black2022gpt, wang2021codet5, chen2021evaluating, li2022competition, nijkamp2022codegen, nijkamp2023codegen2, fried2022incoder, gunasekar2023textbooks, allal2023santacoder}. Similarly, math-focused models, such as Minerva~\citep{lewkowycz2022solving} and MathGLM~\citep{yang2023gpt}, have been developed through specialized training and fine-tuning strategies, including the use of external tools and Chain of Thought techniques~\citep{jelassi2023length,liu2023goat,nye2022show,zhou2022teaching,chen2022program,yang2023gpt,gao2023pal,schick2023toolformer}. 
These models, requiring extensive training, highlight the intensive data demands of LLMs in specialized domains. For example, CodeLlama uses 500 billion tokens for code training, 100 billion tokens for Python training, and more than 20 billion tokens for fine-tuning.


\looseness=-1 \textbf{The Fusion of Large Language Models. } Mixture-of-Experts (MoE) is the neural architecture that distributes tasks among multiple specialized networks (experts) and determines their responsibilities via a gating network~\citep{jacobs1991adaptive}. 
MoE enhances the capabilities of LLMs and has been extensively utilized~\citep{clark2022unified,lou2021cross,kudugunta2021beyond,lepikhin2020gshard,mustafa2022multimodal,zhou2022mixture,riquelme2021scaling,shen2023scaling,jiang2023llm,wan2024knowledge,jiang2024mixtral}. Many studies have endeavored to comprehend the Mixture-of-Experts (MoE) from the perspective of computational cost, with a specific focus on its sparse nature~\citep{shazeer2016outrageously,zoph2022st,zuo2021taming,du2022glam,fedus2022switch,komatsuzaki2023sparse,shen2023mixtureofexperts}. The prevailing belief is that the MoE approach can scale up model parameters without incurring an escalation in computational expense. Some work suggests that experts do not necessarily have distinct expertise~\cite{jiang2024mixtral}, while other work verifies the effectiveness of expert specialization~\cite{dai2024deepseekmoe}. We believe both ways could achieve promising performance, unlike those that train MoE models from scratch, this paper seeks to fuse highly specialized models in the fine-tuning phase. 
Compared to methods like knowledge distillation and knowledge fusion~\cite{wan2024knowledge}, our approach aims to achieve optimal performance by retaining the specialized models and learning to fuse the expertise directly, avoiding potential performance loss brought by inaccurate fashion weight estimation and further distillation training.  


\vspace{-0.1cm}
\section{Our Approach}

Compared to methods like Mixture-of-Experts~\cite{shazeer2016outrageously}, which expands the inner model structure to develop different expertise implicitly during training, our approach focuses on fusing specialist models explicitly aligned with different skill sets at the output level directly.
This section first describes the constitution of the proposed model, \modelname, and then introduces the construction of a mixed instruction tuning dataset, \dataname. 

\vspace{-0.1cm}
\subsection{Model}


The proposed fused model consists of three specialized models (termed as specialists), collectively denoted as $\mathcal{M}_\Theta = \{{E}_\text{text}, {E}_\text{code}, {E}_\text{math}\}$, where ${E}_\text{text}$ is mainly trained on natural language text, ${E}_\text{code}$ is trained on programming code, and ${E}_\text{math}$ is trained on mathematical problems. 
Each specialist model is essentially a native autoregressive large language model. They share the same architectural framework and vocabulary space but are trained on distinct datasets that are representative of their expertise.

\begin{figure*}
    \centering
    \includegraphics[width=0.98\textwidth]{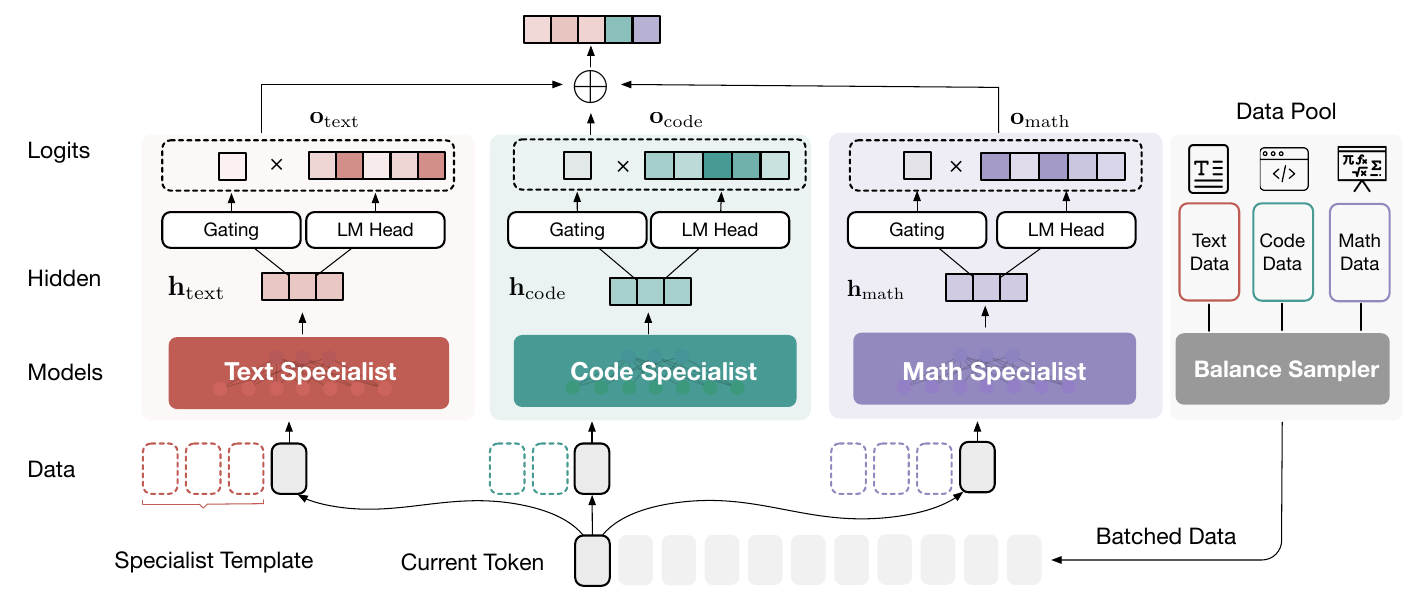}
    \caption{Architectrue of our proposed \modelname framework. We do not show the two-stage training in this illustration.}
    
    \label{fig:moe}
\end{figure*}

\textbf{Architecture.} The fused model aims to utilize the expertise of each specialist model appropriately based on the nature of the input data.
The integration of specialized ability is realized by a shared gating layer $g$ that calculates the weight for each token per specialist. 
Specifically, during training, for the token $x^{(i)}$ concerned, the three specialists output token hidden states $\mathbf{h}^{(i)} =\{\mathbf{h}_\text{text}^{(i)}, \mathbf{h}_\text{code}^{(i)}, \mathbf{h}_\text{math}^{(i)}\}$ and corresponding logits $\mathbf{o}^{(i)} =\{\mathbf{o}_\text{text}^{(i)}, \mathbf{o}_\text{code}^{(i)}, \mathbf{o}_\text{math}^{(i)}\}$ as a native language model. Then, the gating layer $g_\Phi$ is applied to each set of specialist outputs to obtain the final logits.

Practically, the gating layer is implemented as a linear network that calculates the weight for each token $x^{(i)}$ based on the last hidden states $\mathbf{h}^{(i)}=E(x^{(1:i-1)})$. For each token $x^{(i)}$, the final output logits from the fused model are computed as:
\begin{equation}
\begin{split}
    &g_\Phi(\mathcal{M}_\Theta(x^{(i)})) = \mathbf{w}^{(i)T} (\mathbf{o}^{(i)}_\text{text}: \mathbf{o}^{(i)}_\text{code}: \mathbf{o}^{(i)}_\text{math}), \\
    &\text{where}\ \mathbf{w}^{(i)}=\text{Softmax}(g(\mathbf{h}^{(i)}_\text{text}): g(\mathbf{h}^{(i)}_\text{code}): g(\mathbf{h}^{(i)}_\text{math}))
\end{split}
\end{equation}
\textbf{Training.} 
One possible approach to training the model is to train the gating network only, expecting it to allocate each token to its optimal distribution over the three specialists. Such training strategy highly relies on the gating module's capacity in capturing the complex and diverse context in drastically different instructions. An easier way to boost the performance is to jointly fine-tune the three specialists along with the gating module. However, the specialists can be negatively impacted by gradients back-propagated from the gating module due to its poor performance at the early stage, which may cause irreversible damage to the specialist's inherent ability.

\looseness=-1 To tackle the problem, we propose a two-stage training strategy to ensure training stability and mitigate potential specialist ability loss. The first stage trains only the gating module parameters for $N_1$ steps and keeps specialists frozen. The purpose is to retain specialist capability while warming up the gating module. After the first stage of training, the gating network is expected to output reasonable token weights that favor over specific specialists according to data type.
The second stage continues to fine-tune all model parameters based on the first stage for $N_2$ steps. At this stage, the specialist models are jointly optimized for a better overall performance.
At both stages, the training loss is the cross-entropy loss given true labels $y$ and the final model output.


\begin{equation}
\begin{aligned}
    \mathcal{L}(x, y) &= \sum_{i} \mathcal{CE}(g_\Phi(\mathcal{M}_\Theta(x^{(i)})), y^{(i)}).
\end{aligned}
\end{equation}

The training proceeds by minimizing the total loss over all instances in the training set using a suitable optimization algorithm, such as AdamW. 
The gradients are back-propagated through both the specialist models and the gating networks, allowing the gating mechanism to learn how to distribute the inputs effectively among the specialists. The overall training process is shown in Algorithm~\ref{alg:train}.

\textbf{Data-level Balancing.} 
Since all specialists are well aligned to one specific type of instruction, they may demonstrate different activation patterns that are highly sensitive to inputs.
Therefore, to fully take advantage of the specialized ability, we use specialist-specific templates to format our training data (see Appendix~\ref{appendix:details}). Each training sample is wrapped up by three different templates and fed into the respective specialist model. Since the loss is only calculated for the model response part, the response tokens will still be aligned, and their logits can be fused together seamlessly.
We also adopt a batch-level class-balance sampler during training. 
The sampler ensures that each training batch contains the same number of training instances from the three categories, thus ensuring that the three specialists are activated and optimized at similar level for each batch, preventing from biased training that favor over one specific specialist. 
As shown in Algorithm~\ref{alg:train}, each batch of data contains $n \times 3$ instances in total.
We explain the reason to alleviate the imbalance issue in the data-level and validate the effectiveness of the class-balance sampler in Section~\ref{sec:ablation}.
\begin{algorithm}[tb]
   \caption{Algorithm for two-stage training with balanced data sampler, where $\mathcal{S}(\mathcal{D}, n)$ means randomly sampling $n$ examples from dataset $\mathcal{D}$. $N_1$ and $N_2$ are total training steps, and $\eta_1$ and $\eta_2$ are the scheduled learning rate for the two stages, respectively.}
   \label{alg:train}
\begin{algorithmic}
   \STATE {\bfseries Input:} specialized models $\mathcal{M}_\Theta$, gating $g_\Phi$, training data $\mathcal{D}_{\text{text}}, \mathcal{D}_{\text{code}}, \mathcal{D}_{\text{math}}$
   \FOR{$i=1$ {\bfseries to} $N_1$}
        \STATE $\mathcal{D}^i = \bigcup_{t \in \{\text{text}, \text{code}, \text{math}\}} \mathcal{D}^i_t$ \\
        $\ \ \ \ \ \ = \bigcup_{t \in \{\text{text}, \text{code}, \text{math}\}}\mathcal{S}(\mathcal{D}_{t}, n)$
        \STATE $g_\Phi = g_\Phi -  \eta_1 \Delta_\Phi \frac{1}{|\mathcal{D}^i|}\sum_{(x,y)\in\mathcal{D}^i}\mathcal{L}(x, y)$
   \ENDFOR
   \FOR{$j=1$ {\bfseries to} $N_2$}
        \STATE $\mathcal{D}^j = \bigcup_{t \in \{\text{text}, \text{code}, \text{math}\}} \mathcal{D}^j_t$ \\
        $\ \ \ \ \ \ = \bigcup_{t \in \{\text{text}, \text{code}, \text{math}\}}\mathcal{S}(\mathcal{D}_{t}, n)$
        \STATE $g_\Phi = g_\Phi -  \eta_2 \Delta_\Phi \frac{1}{|\mathcal{D}^j|}\sum_{(x,y)\in\mathcal{D}^j}\mathcal{L}(x, y)$
        \STATE $\mathcal{M}_\Theta = \mathcal{M}_\Theta -  \eta_2 \Delta_\Theta \frac{1}{|\mathcal{D}^j|}\sum_{(x,y)\in\mathcal{D}^j}\mathcal{L}(x, y)$
   \ENDFOR
\end{algorithmic}
\end{algorithm}

\textbf{Inference.} The model design adopts post-specialist token-level gating, meaning that all specialists are activated during inference. For each token $x^{(i)}$, the three specialist models $\mathcal{M}_\Theta$ are queried, and their logits are fused using the gating module $g_\phi(\cdot)$ as in the training phase. The softmax is applied to the aggregated logits to generate probabilities for the next token. The selected token is then used as part of the input for the subsequent inference step in an autoregressive manner. Our design opens doors for sophisticated, real-time adaptability that monolithic models lack.
For example, in a text string interwoven with mathematical equations and code snippets—common in scientific papers, the fused model can shift its ``attention'' between specialists within the same sequence, ensuring that each token is treated with the most appropriate domain expertise.
But on the other hand, since all specialists are activated in inference, computational overheads are inevitably introduced. 
In experiments, we adapt the vLLM project \cite{kwon2023efficient} to our fused model to accelerate inference, which is elaborated in Appendix~\ref{appendix:vllms}.

\textbf{Why Not Sample-Level?} One direct and simple approach to fusing specialized models is to train them in a sample-level manner.
That is, freezing the specialist models and directly train a selector, letting one specialist respond to a whole query. 
This approach seems to safeguard the lower-bound performance for the model effectively, so why does this paper opt for token-level training rather than sample-level? The main reason is that, although this paper categorizes the data into three distinct symbolic systems, they may blend together in real-world queries (for instance, code data may contain extensive text intended for documentation). Similarly, while these three capabilities might weaken each other in some respects, they could also enhance one another in different contexts, which is demonstrated in Section~\ref{sec:results}. 
We choose to design the fused model to seek a higher performance ceiling.




\begin{table*}[ht]
\centering
\caption{Statistics and information of \dataname \  dataset. \# Topics are the number of meta-topics and sub-topics.}
\scalebox{0.85}{
\begin{tabular}{p{1.3cm}|ccc}
\toprule
          & \textbf{Text Part}   & \textbf{Code Part}   & \textbf{Math Part}   \\ \midrule
\# Data        & \cellcolor{mycolor1}100,000 & \cellcolor{mycolor2}100,000 & \cellcolor{mycolor3}110,000 \\
\# Topics &   \cellcolor{mycolor1}30/1100     & \cellcolor{mycolor2}21/407    & \cellcolor{mycolor3}21/80     \\ 
Examples    &  \begin{minipage}{0.32\textwidth}
      \includegraphics[width=\textwidth]{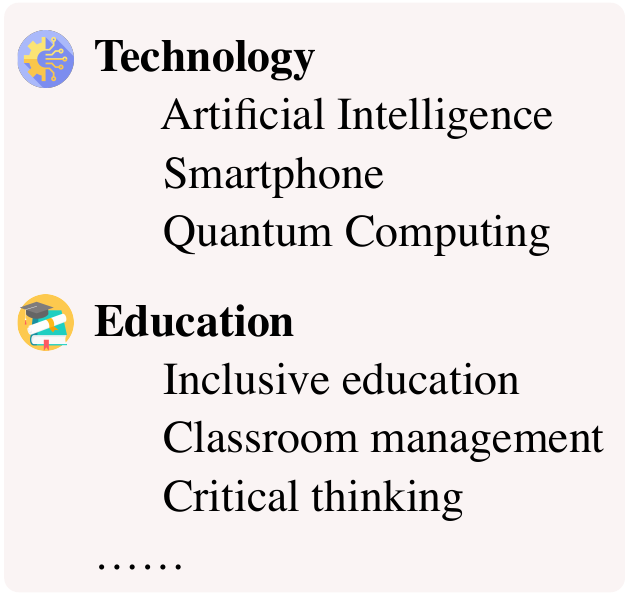}
    \end{minipage}    & \begin{minipage}{0.32\textwidth}
      \includegraphics[width=\textwidth]{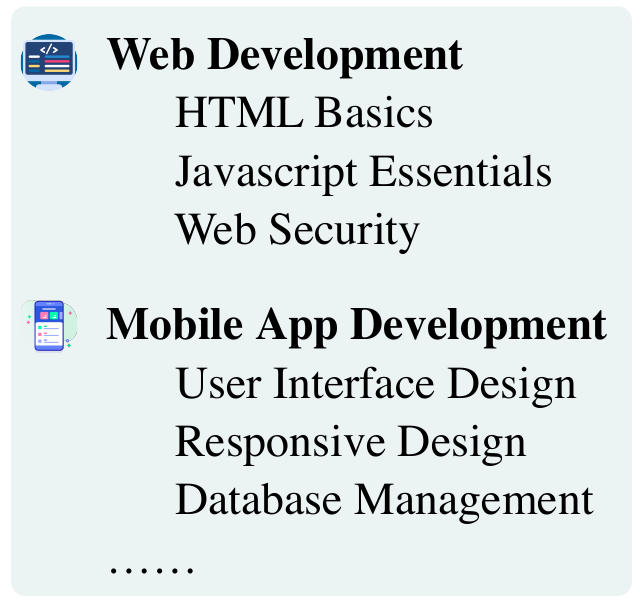}
    \end{minipage}       &  \begin{minipage}{0.32\textwidth}
      \includegraphics[width=\textwidth]{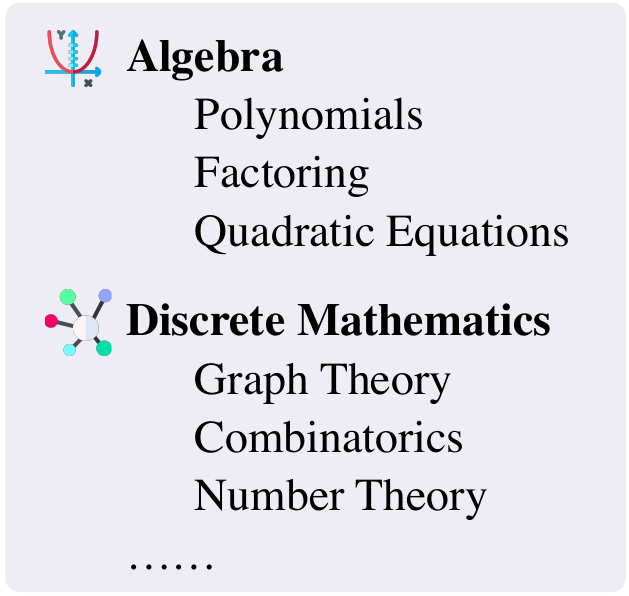}
    \end{minipage}     \\ \bottomrule
\end{tabular}}
\end{table*}

\subsection{\dataname \ Dataset}

Currently, within the open-source community, there are already multiple instruction-tuning datasets for text-based conversations. However, there is a relatively limited amount of systematic code and mathematical instruction tuning data available.
In this section, we construct \dataname, a comprehensive dataset tailored for training our proposed model. 
\dataname \ spans a wide range of subject matter, covering natural language, coding, and mathematical instructions.
We employ a multi-stage pipeline to generate a rich set of instructional data.
First, we engage in multi-turn interactions with GPT-4, constructing meta-topics that best represent each domain. Then, each meta-topic is utilized to generate multiple sub-topics. 
For each sub-topic, LLM is tasked with generating diverse and informative specific instructions. After obtaining these instructions, we continue with in-context learning, generating both strong and weakly related instructions for each directive to fully leverage LLM's generalization capabilities. Finally, we extract 30\% of the instruction data and make them more complex. Once we have the complete pool of instructions, we use GPT-4 to respond to these instructions, resulting in \dataname.
\begin{figure}
    \centering
    \includegraphics[width=0.46\textwidth]{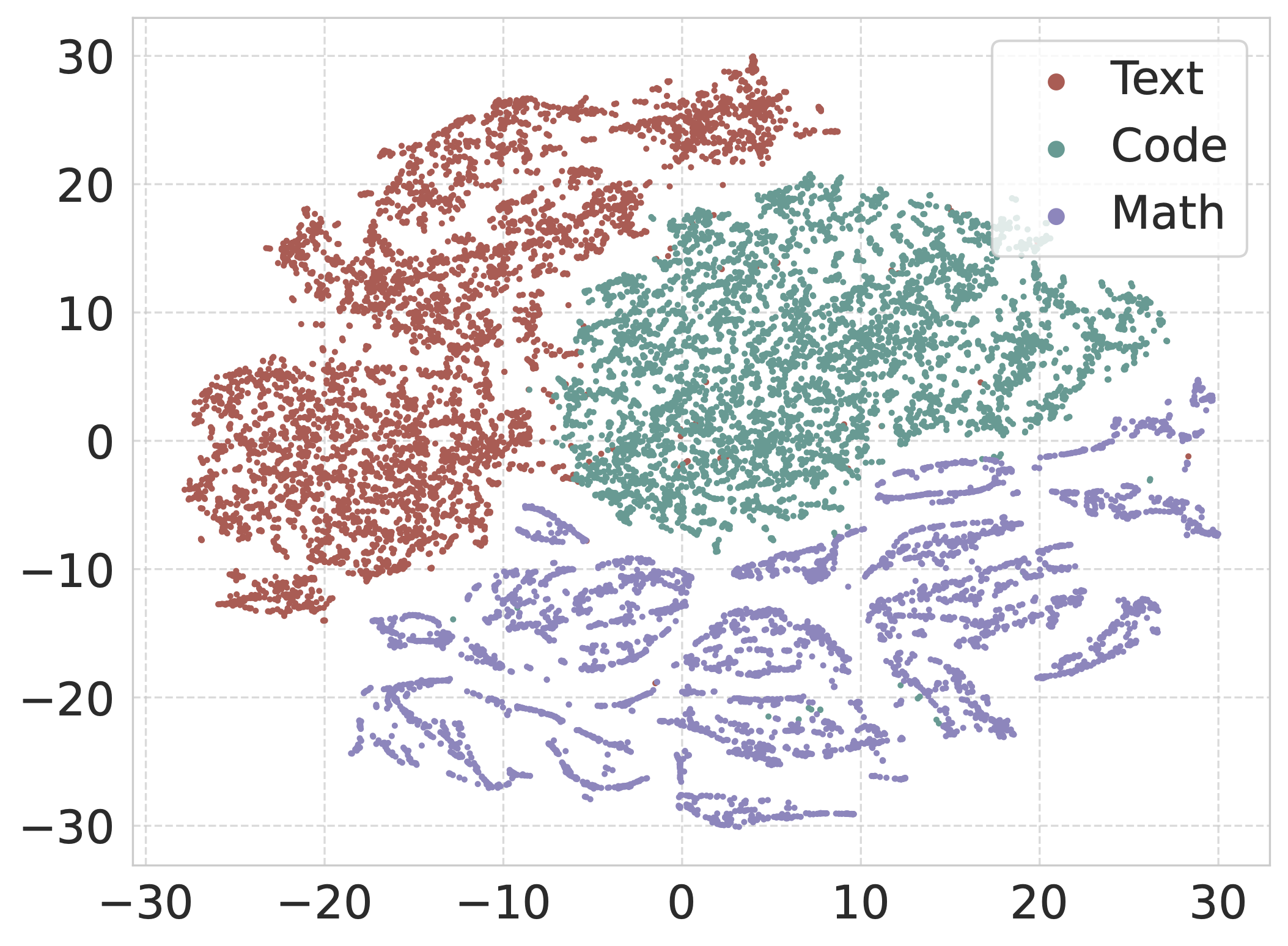}
    \caption{t-SNE visualization of \dataname \ dataset.}
    \vspace{-0.4cm}
    \label{fig:mix_vis}
\end{figure}

\begin{figure}[!th]
    \centering
    \includegraphics[width=0.43\textwidth]{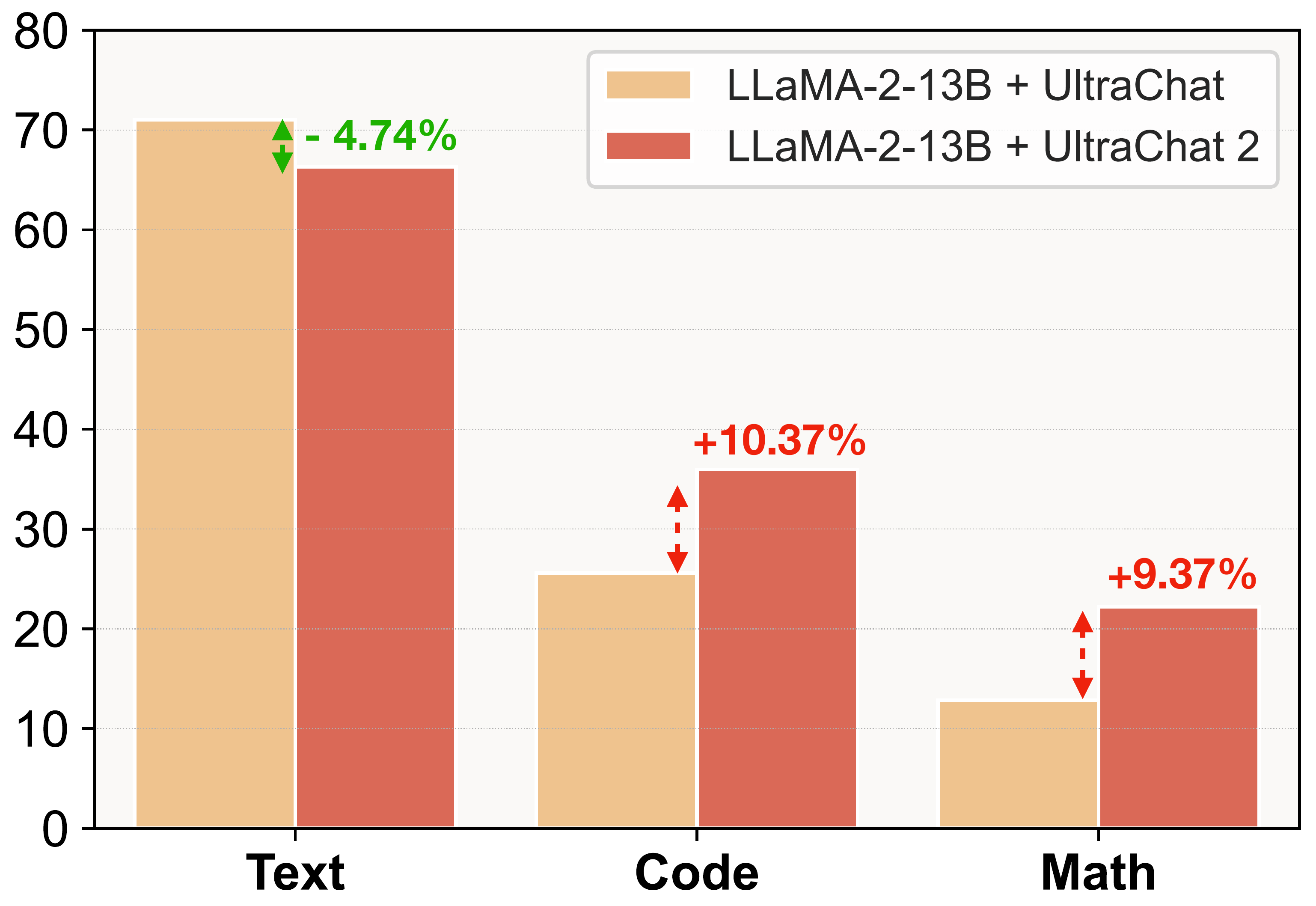}
    \caption{Performance comparison of Llama-2 model trained on \textsc{UltraChat} and \dataname.  The performance for the text domain is computed by the average results on TruthfulQA (Acc) and AlpacaEval (Win Rate) datasets; the performance for the code domain is Pass@1 of HumanEval; and the performance for the math domain is the average result of GSM8K (Pass@1) and MATH (Pass@1). 
    }
    \vspace{-0.3cm}
    \label{fig:comp}
\end{figure}

\textbf{Data Analysis.} We randomly sample 5000 instructions from each category and 
visualize the data distribution in in Figure~\ref{fig:mix_vis}. The representations are obtained by averaging the last layer of hidden states of each token from Llama-2-13B, and dimensions are further reduced by the t-SNE algorithm~\cite{van2008visualizing}. The visualization clearly demonstrates the diversity and distinctiveness of different types of \dataname, which aligns with the intuition and echos the discussion in Section~\ref{sec:intro}. 
\dataname \ provides high-quality resources for the facilitation of specialized models. 
\looseness=-1 We train a Llama-2-13B on \dataname~to give a glance at the effectiveness.
As shown in Figure~\ref{fig:comp}, in the text domain, the Llama-2-13B + UltraChat 2 configuration exhibits a 3.9\% decrement in performance relative to the baseline that is only trained on the text domain. 
Conversely, in the code domain, there is a significant performance increment of 10.4\% with the UltraChat 2 enhancement.  The math domain also shows a performance increase of 9.4\% with the UltraChat 2 integration, indicating a clear advantage of the updated system in code-related and mathematical reasoning tasks.

\section{Experiments}
\looseness=-1 We conduct extensive experiments to analyze the effectiveness and behaviors of \modelname. Implementation details are reported in Appendix~\ref{appendix:details}.
Our models, data, training and inference frameworks will be publicly released.

\subsection{Experimental Settings}

\looseness=-1 \textbf{Models.} To validate the effectiveness of our approach,
we adopt Llama-2-13B~\cite{touvron2023llama} as the backbone for experiments.
Specifically, we use UltraLM-13B-V2.0~\cite{ding2023enhancing}, CodeLlama-13B-instruct~\cite{roziere2023code}, WizardMath-13B-V1.0~\cite{luo2023wizardmath} as the three specialist models. 
All model parameters are fine-tuned under the proposed \modelname\  framework.

\looseness=-1 \textbf{Evaluation.} For the text domain, we use TruthfulQA~\cite{lin2021truthfulqa} and AlpacaEval~\cite{alpaca_eval} for evaluation. The former is more focused on the truthfulness of LLMs, and the latter consists of more general natural language questions.
For the code domain, we use HumanEval~\cite{chen2021evaluating} for evaluation, which is a code completion task. 
For the math domain, we use GSM8K~\cite{cobbe2021training}, MATH~\cite{hendrycks2020measuring}, SAT-MATH~\cite{zhong2023agieval} and AQuA~\cite{ling-etal-2017-program} for evaluation.
For evaluation, we transform each dataset into instruction format, and use consistent template as training for inference. 
Specifically, we evaluate under the MC2 setting in TruthfulQA, where each option is fed into the model independently and the model is queried for true or false judgment. For HumanEval, we use InstructHumanEval that transforms the original dataset into instruction format.
All results are zero-shot and produced by our experiments. We do not use any chain-of-thought (CoT) techniques to boost the performance.

\begin{table*}[!th]
\centering
\caption{Results of baselines and our proposed models across different benchmarks. \textbf{All the numbers are {zero-shot} results} produced by our experiments under the same inference framework. \textit{No} Chain-of-thought (CoT) techniques are employed in evaluation. The highest results are \textbf{bold}, and the second highest results are \underline{underlined}.
Delta values in the second block mean the performance differences between the model and the corresponding specialist models. 
}
\scalebox{0.86}{
\begin{tabular}{lcccccccccc}
\toprule 
\multirow{2}{*}{\textbf{Model}} &  \cellcolor{mycolor1} \textbf{TruthfulQA} &  \cellcolor{mycolor1} \textbf{AlpacaEval}  &  \cellcolor{mycolor2} \textbf{HumanEval} & \cellcolor{mycolor3} \textbf{GSM8K} & \cellcolor{mycolor3}  \textbf{MATH} & \cellcolor{mycolor3}  \textbf{SAT-Math} & \cellcolor{mycolor3}  \textbf{AQuA} & \textbf{Avg.} \\  
& \cellcolor{mycolor1} Acc (\%) & \cellcolor{mycolor1} Win Rate (\%)  & \cellcolor{mycolor2} Pass@1 (\%)   & \cellcolor{mycolor3}  Pass@1 (\%) & \cellcolor{mycolor3} 
 Pass@1 (\%) & \cellcolor{mycolor3}  Acc (\%) & \cellcolor{mycolor3}  Acc (\%)
\\ 
\midrule
UltraLM-2 &\underline{58.82}  & \textbf{83.23} & 25.61 & 25.09 & 4.48 & 25.00 & 25.98 & 35.46 \\ 
CodeLlama & 56.89  & 69.21 & \underline{48.78} & 23.12 & 6.16 & 27.73 & 25.59 & \underline{36.78} \\ 
WizardMath &26.81  & 51.50 & 10.98 & \textbf{56.18} & \textbf{12.2} & \underline{29.55} & \underline{22.05} & 29.91 \\ \midrule
\modelname & \textbf{64.67}  & \underline{82.35}  & \textbf{53.03} & \underline{54.59} & \underline{11.36} & \textbf{30.00} & \textbf{26.38} & \textbf{47.48} \\ \bottomrule
\end{tabular}}
\label{tab:main}
\end{table*}

\subsection{Results on Benchmarks}
\label{sec:results}
\looseness=-1 Our approach involves further training already highly specialized models, but to what extent can this retraining be effective without the \modelname \  framework? 
We provide two sets of baselines. The first set is the specialists of three domains, and the second group is the further trained versions of such specialists with identical training data as \modelname.
Comparing to the original specialist models as shown in Table~\ref{tab:main}, \modelname\ can consistently produce on-a-par or even superior performance across benchmarks from different domains and achieves the highest overall results. Notably, \modelname\ significantly outperforms respective specialists on TruthfulQA and HumanEval datasets by 5.86\% and 4.25\%, indicating that the three specialist models interact with each other in helpful ways to boost performance on more comprehensive datasets. The result demonstrates the effectiveness of directly fusing specialist models with the proposed framework in both retaining and potentially synthesizing expertise to achieve even better performance.

\begin{figure*}[htbp]    
  \centering            
  \subfloat[Text-specialized models.]   
  {
      \label{fig:subfig1}\includegraphics[width=0.32\textwidth]{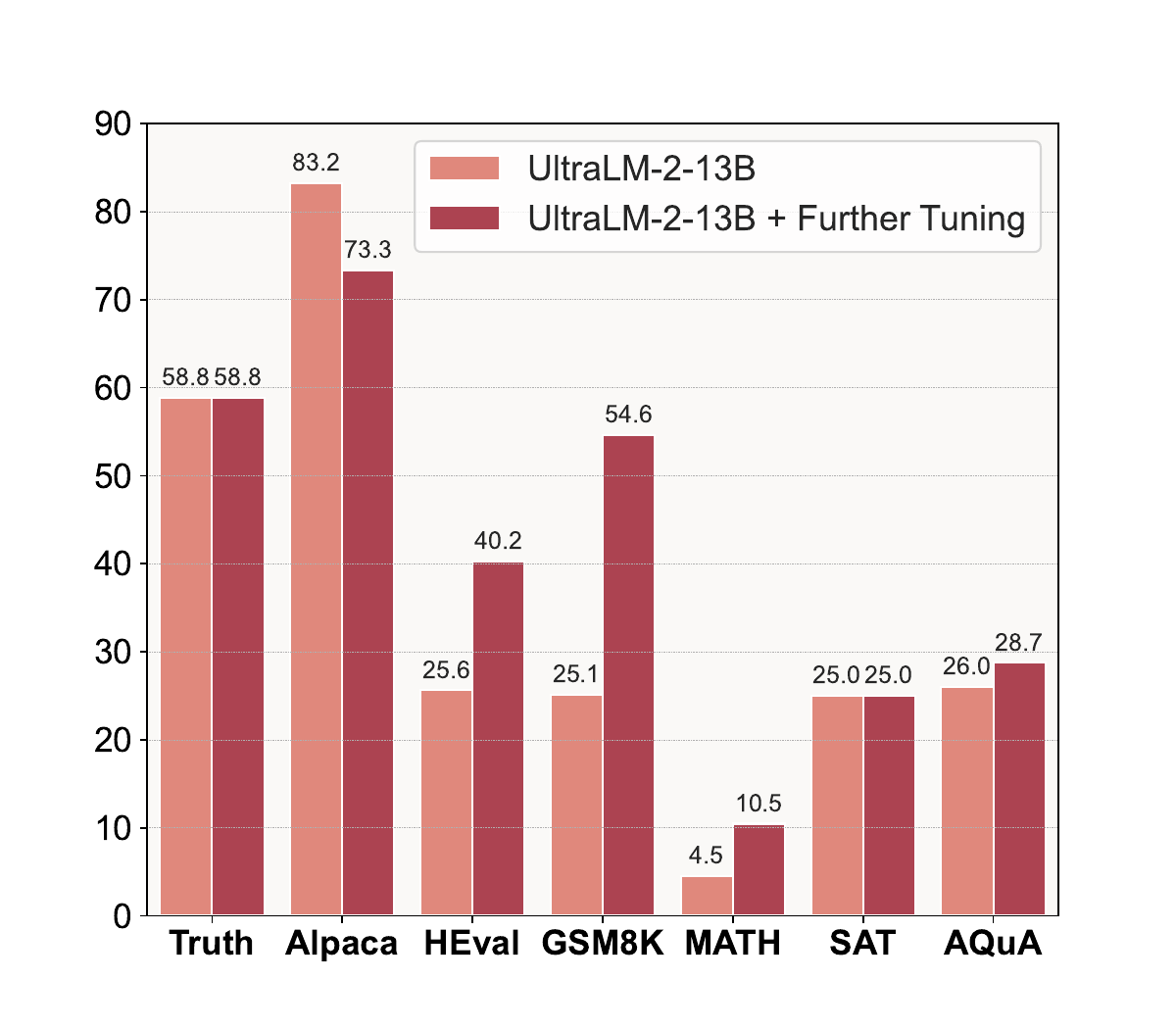}

  }
  \subfloat[Code-specialized models.]
  {
      \label{fig:subfig2}\includegraphics[width=0.32\textwidth]{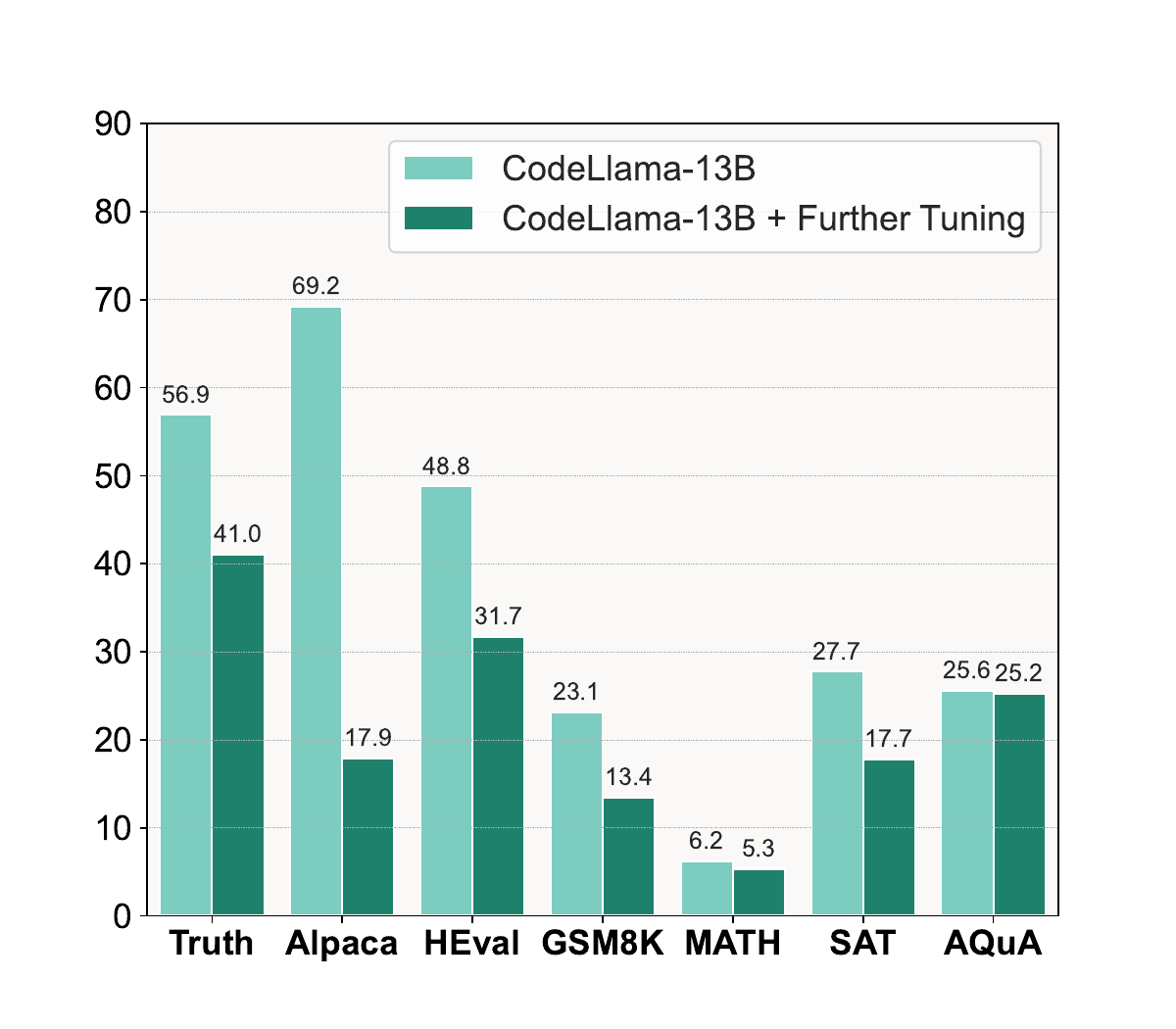}
  }
    \subfloat[Math-specialized models.]
  {
      \label{fig:subfig3}\includegraphics[width=0.32\textwidth]{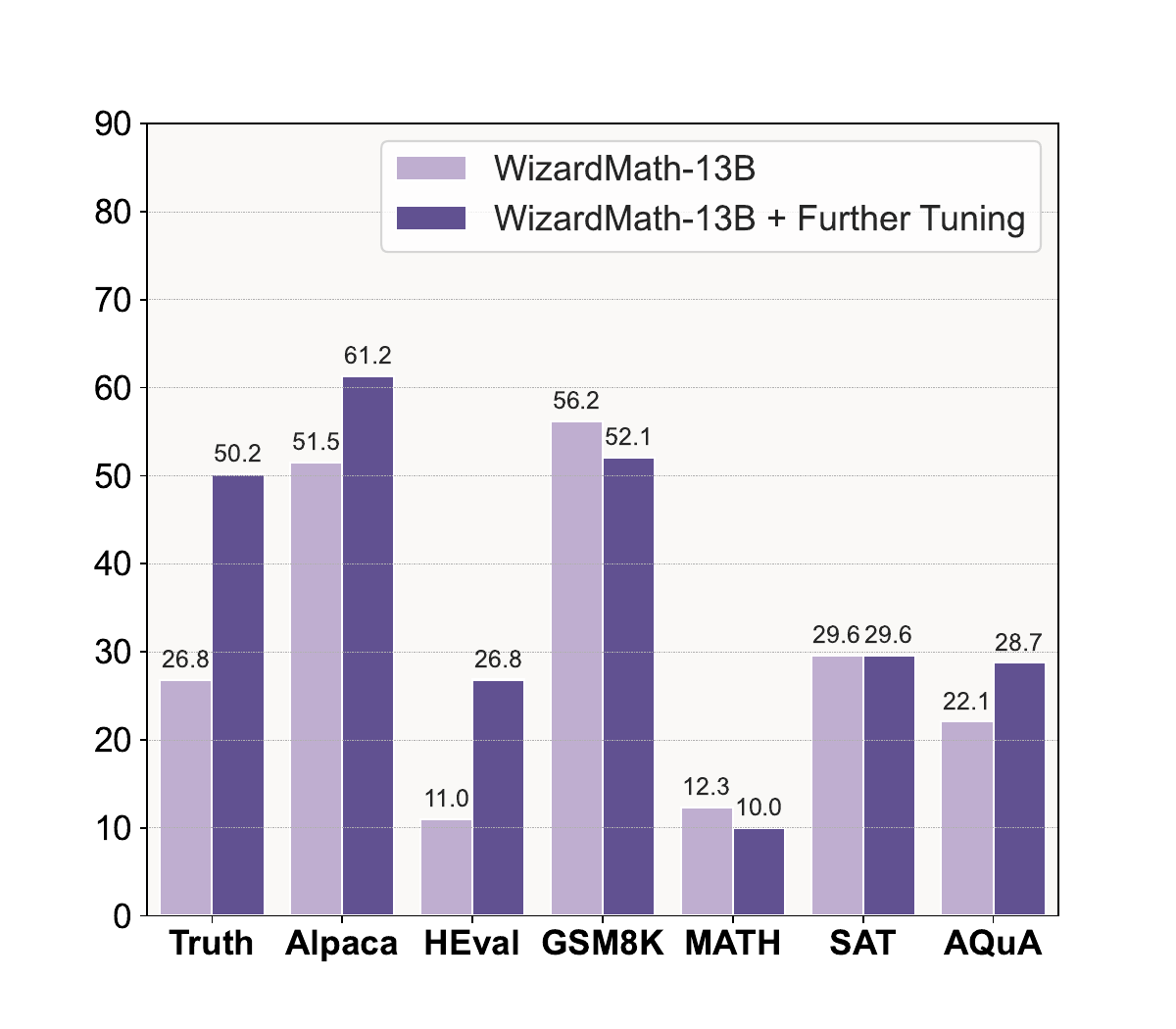}
  }
  \caption{Performance comparisons between specialist models and the further training versions of them.}    
  \label{fig:comp-further}            
  \vspace{-0cm}
\end{figure*}

Furthermore, directly fine-tuning specialist models on our training data may not produce desirable performance, as shown in Figure~\ref{fig:comp-further}. Although training on a mixed dataset indeed boosts other expertise domains, it also severely harms the original expertise of the model concerned, which can be seen from the performance drop after further tuning. 
Meanwhile, it should be aware that the three models not only differ in expertise but also in the training stages they each have come through. Different specialist models show distinct patterns of expertise distribution after further tuning. Models like UltraLM and WizardMath, which are directly instruction-tuned based on Llama, gain more benefit from further tuning. It should be noted that both UltraLM and WIzardMath improve significantly on coding tasks after further tuning (14.6\% and 15.8\%), while UltraLM's performance on GSM8K and MATH doubles after tuning.
However, CodeLlama's performance, unfortunately, degrades on every benchmark, especially in complex instruction following tasks like Alpaca. It is probably because CodeLlama has undergone thorough code infilling pre-training (500B tokens) before instruction fine-tuning. The model parameters may be detrimentally disrupted when directly tuning on drastically different domains of data like pure text and mathematical notation, and therefore suffer from great loss of basic instruction following ability. This points to the fact that the outcome of continuously fine-tuning a well-aligned model highly relies on the previous training data schedule and training strategy adopted, while the proposed framework could seamlessly bridge distinctive model expertise with simple tuning methods and mixed data.
Among the three specialists, UltraLM-2 seems to benefit the most in terms of overall performance after further tuning, indicating that a ``specialist'' in text may be equipped with much broader expertise and may have more potential in expanding new expertise from further supervised fine-tuning. Such phenomenon is intuitively correct as text domain does incorporate much more diversity compared to the other two, but it also sheds light on the fact that there is much more fine-grained variability that should be captured by the term ``expertise''. As we only distinguish the ``expertise'' based on the difference in symbol system, further disentanglement and more experiments are worth exploration. Above all, our experimental results show that \modelname\ can easily produce a well-rounded system by fusing and leveraging expertise from different specialist models.

\subsection{Results on MT-Bench}
Table~\ref{tab:mtbench} shows the results on MT-Bench. Overall, \modelname achieves highest performance across different categories of tasks, demonstrating the effectiveness of our proposed expert fusing methods compared to direct further tuning on domain specialized model. It can also be seen that UltraLM and its further-tuned version still have advantage on some tasks like information extraction and STEM-related problem solving, while the code and math counterparts face significant performance drop after further tuning. The results indicate the better generalizability of a text specialized model over code and math domains in terms of direct further tuning, while fusing them together would yield a more suitable model structure for comprehensive future enhancement.
\begin{table*}[!ht]
    \centering
    \scalebox{0.85}{
    \begin{tabular}{l|ccccccccc}
        \toprule
        \textbf{Model} & \textbf{Writing} & \textbf{Roleplay} & \textbf{Reasoning} & \textbf{Math} &\textbf{ Coding} & \textbf{Extraction} & \textbf{STEM} & \textbf{Humanities} & \textbf{Overall} \\
        \midrule
        UltraLM-2 & \textbf{8.83} & \underline{7.98} & \textbf{5.20} & 2.90 & 4.00 & \textbf{6.74} & 8.08 & \textbf{9.46} & \underline{6.62} \\
        CodeLlama & 5.80 & 7.10 & 3.80 & 3.05 & 3.43 & 5.36 & 5.65 & 7.05 & 5.16 \\
        WizardMath & 7.75 & 7.03 & 4.80 & 3.85 & 3.50 & 4.65 & 7.65 & 9.13 & 6.04 \\
        \midrule
        UltraLM-2+Further Tune & 7.85 & 7.60  & 4.30 & \underline{4.48} & \textbf{5.20} & 5.78 & \textbf{8.32} & 9.40 & \underline{6.62} \\
        CodeLlama+Further Tune & 7.33 & 6.60 & 3.85 & 2.25 & 3.68 & 5.20 & 4.68 & 5.10 & 4.84 \\
        WizardMath+Further Tune & 7.18 & 6.90 & 4.95 & 4.25 & 4.55 & 5.18 & 7.55 & 7.98  & 6.07 \\ \midrule
        \modelname & \underline{8.60} & \textbf{8.11} & \underline{5.00} & \textbf{5.15} & \underline{5.10} & \underline{6.53} & \underline{8.23} & \underline{9.43} & \textbf{7.02} \\
        \bottomrule
    \end{tabular}}
    \caption{Performance on MT-Bench. The highest results are \textbf{bold}, and the second highest results are \underline{underlined}.}
    \label{tab:mtbench}
\end{table*}

\vspace{-0.2cm}
\subsection{Ablation Study}
\label{sec:ablation}

Training fused models could cause load imbalance, leading to the collapse of the routing mechanism.
A typical approach to mitigate this issue in MoE is to introduce a balance loss to prevent certain models from being over-selected or under-selected. 
In our framework, we do not introduce explicit balance loss based on a simple hypothesis: A model that has been highly specialized can automatically produce a lower loss on the data it is good at. 
Now that the model already has data that is good at processing, we hope to solve the problem from the data level, not force the specialist models to participate to a certain extent during the calculation.
We find that designing some training methods can make the progress more stable. 
Two key components of our framework are \textit{two-stage training} and \textit{balanced sampler}.
The former plays a role similar to warm-up, allowing the randomly initialized gating module to adapt to the current expert model. The latter, as mentioned, ensures load balance at the data level. In Table~\ref{tab:ablation}, we report the best performance under each training strategy. It can be observed that the beneficial effects of these two modules are obvious, and their use has made the overall performance of the fused model improve considerably. We further investigate the impact on the training stability of the balance sampler. We train two versions of the model with the same dataset and sample 12 checkpoints, respectively, from 2000 steps to 9000 steps, and conduct evaluations. As shown in ~\ref{tab:stable}, with the help of the balance sampler, the fused model could achieve superior performance and lower standard deviations on all datasets. GSM8K is relatively stable during training, however HumanEval may face larger fluctuations.

\begin{table}[!ht]
    \caption{Results across TruthfulQA (Truth), HumanEval (H-Eval), and GSM8K with different training strategies.}
    \centering
    \scalebox{0.9}{
    \begin{tabular}{lccc}
    \toprule
     \textbf{Strategy} &  \textbf{Truth} & \textbf{H-Eval} & \textbf{GSM8K} \\ \midrule
       Direct Training  & 51.17 &  46.95 & \underline{53.83} \\
       + Two-Stage &  \underline{61.72} & \underline{50.00} & 52.69 \\
       {+ Two-Stage + Balanced}& \textbf{64.67} & \textbf{53.05} & \textbf{54.59} \\
    \bottomrule
    \end{tabular}}
    \vspace{-0.3cm}
    \label{tab:ablation}
\end{table}

\begin{table}[!ht]
    \caption{Mean results and standard deviation over 12 checkpoints with and without the balance sampler (two-stage training are both applied).}
    \centering
    \scalebox{0.9}{
    \begin{tabular}{lccc}
    \toprule
     \textbf{Strategy} &  \textbf{Truth} & \textbf{H-Eval} & \textbf{GSM8K} \\ \midrule
     w/o Balance  & 57.54±2.80 &  48.27±4.92 & 52.91±1.76 \\
       w/ Balance &  \textbf{59.91±1.96} & \textbf{{53.68}±2.74}   & \textbf{{53.77}±1.74}\\
    \bottomrule
    \end{tabular}}
      \vspace{-0.3cm}
    \label{tab:stable}
\end{table}

\begin{table}[!ht]
    \caption{Average weights from three specialist models of different data.}
    \centering
    \scalebox{0.9}{
    \begin{tabular}{lccc} \toprule
     \textbf{Avg. Weight}   & \textbf{Text Data} & \textbf{Code Data} & \textbf{Math Data} \\ \midrule
      $\mathbf{w}_\text{text}$ & \textbf{0.45}& 0.23 & 0.18\\
      $\mathbf{w}_\text{code}$ & 0.29  & \textbf{0.59} & 0.39 \\
      $\mathbf{w}_\text{math}$ & 0.26 & 0.18 & \textbf{0.43} \\ \bottomrule
    \end{tabular}}
    \vspace{-0.4cm}
    \label{tab:weight}
\end{table}

\begin{figure*}[!ht]    
  \centering            
  \subfloat[Weights distribution of three specialist models for 100 text data samples.]   
  {
      \label{fig:subfig1}\includegraphics[width=0.99\textwidth]{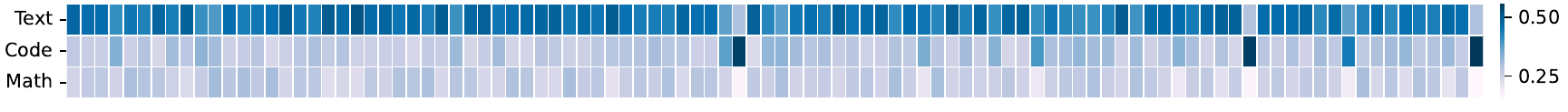}

  }
  \\
  \subfloat[Weights distribution of three specialist models for 100 code data samples.]
  {
      \label{fig:subfig2}\includegraphics[width=0.99\textwidth]{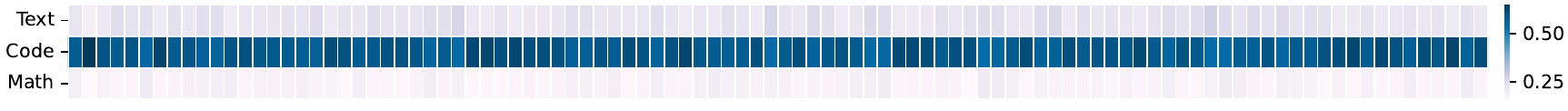}
  }
  \\
    \subfloat[Weights distribution of three specialist models for 100 math data samples.]
  {
      \label{fig:subfig3}\includegraphics[width=0.99\textwidth]{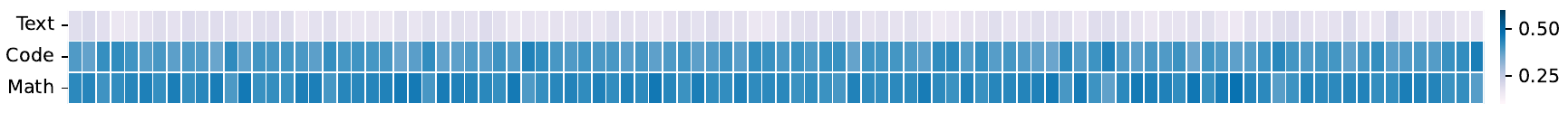}
  }
  \caption{Weight distributions of 300 data samples from text, code, and math domains. Each column is a data point, and each row is the average weight of one specialist model. The darker the color, the more average weight the model gives to the tokens of this data point.}  
  \vspace{0cm}
  \label{fig:weight}            
\end{figure*}

\subsection{Expertise Analysis}
\label{sec:analysis}

In our training, there is no explicit mechanism to make certain specialists ``pay more attention'' to the corresponding data. But as mentioned, we expect that the specializations could still be separated, and a type of data will receive different gating weights from the specialist models. We randomly sample 100 data instances from the three domains and conduct analysis by directly going through the inference to the fused model, and calculate the weight from three specialist models of each token. 
Table~\ref{tab:weight} shows the average weights of all the tokens in each set of data from three specialist models. And intuitively, each set of data is primarily driven by the corresponding specialist model. The prominence of code data is evident, with the corresponding expert models significantly outweighing the other two models. In mathematical data, code and mathematical models almost equally dominate inference, with a marginal difference. This is more distinctly observable in the sample-level distribution illustrated in Figure~\ref{fig:weight}. 
Despite the fusion and further training of the models, it's evident that these specialized models still retain their original functionalities and are now capable of synergistic performance.



\section{Conclusion}
This paper aims to integrate coding and mathematical reasoning capabilities into a general language model with as little loss as possible. We present \modelname, a simple framework to train high-specialized models with a token-level gating mechanism and a two-stage balanced training strategy. Accompanied by the goal, we construct a high-quality and diverse instruction tuning dataset, \dataname, that contains 300,000 instructions and responses from 3 domains, 72 meta-topics, and 1587 sub-topics. Our experiments demonstrate the effectiveness of the proposed framework by showing that fused models can be performative simultaneously in text understanding, code generation, and mathematical reasoning. In future work, the proposed \modelname can also be adapted to domains beyond the mentioned ones. For example, by using to fuse language models that are specialized in different languages.

\section*{Broader Impact}
This work aims to integrate professional code generation and mathematical reasoning capabilities into a general interactive language model, proposing a framework to fuse highly specialized models and construct a corresponding dataset. 
The goal of the study is to further advance and leverage the research of open-source large language models, providing a potential solution to bring together specialized language models into an integrated and comprehensive system.
The study, therefore, shares many of the potential impacts with other research on LLMs, e.g., hallucination and biased output. The model outputs should be closely supervised before put into use. 
On the other hand, this work sheds light on the possibility of fusing language models that are trained on drastically different datasets and may benefit future applications of large language models. How to better define and choose ``expertise'' is also worth exploration under the framework of \modelname.
As for \dataname, we do not use any real human queries or responses to protect privacy. In our manual assessment, the code and math parts of the data are mostly unbiased, but there may still be biased statements in the text part.

\bibliography{example_paper}
\bibliographystyle{icml2024}


\newpage
\appendix
\onecolumn

\section{Implementation Details}
\label{appendix:details}

The gating module is implemented as a two-layer linear model with ReLU~\cite{Agarap2018DeepLU} activation in between. The hidden size of the module is set according to the hidden size of the specialized models.
The gating layer is trained for $N_1=400$ steps at the first training stage with sample size $n=64$ for all experiments and learning rate $\eta_1=2e-5$ is used with a cosine scheduler.
For the second stage with Llama backbone, we use $\eta_2=2e-5$, sample size $n=32$ with cosine scheduler.
Note that our framework requires the consistent tokenization strategy across all specialist models. Therefore, we use the original LLaMA-2-13B tokenizer for \modelname\ training.
All experiments are conducted on $8 \times$ 80GB A100 GPUs and use AdamW optimizer~\cite{Loshchilov2017DecoupledWD}.
Apart from the curated \dataname, we also employ extra instruction tuning datasets from both math and code domains to enrich instructional format diversity. 
Specifically, we use the Evol-Instruct dataset~\cite{luo2023wizardcoder,luo2023wizardmath} for programming and the MathInstruct training set~\cite{yue2023mammoth} for math problems. 
We conduct comprehensive search and filtering (13 grams) to avoid data contamination.

Table~\ref{tab:template1} and Table~\ref{tab:template2} show the conversation templates we use for each specific specialist model and the prompt for converting datasets to instructions in evaluation. In training, each example is wrapped by three different conversation templates and fed into the respective model. In inference, before applying the conversation template, dataset-specific prompt is used to wrap the example first (if applicable).
\begin{table}[!ht]
    \centering
    \scalebox{0.97}{
    \begin{tabular}{l|l}
    \toprule
        \textbf{Model} & \textbf{Conversation Template} \\ \midrule
        UltraLM-2 & \texttt{User: \{instruction\}\textbackslash nAssistant: }\\ \midrule
        CodeLlama &  \texttt{\textless s\textgreater [INST] \{instruction\} [/INST]} \\ \midrule
        WizardMath & \makecell[l]{\texttt{Below is an instruction that describes a task. Write a response that }\\\texttt{appropriately completes the request.\textbackslash n\textbackslash n} \\\texttt{\#\#\# Instruction:\textbackslash n\{instruction\}\textbackslash n\textbackslash n\#\#\# Response:}} \\
    \bottomrule
    \end{tabular}}
    \caption{Model-specific conversation templates for training and evaluation.}
    \label{tab:template1}
\end{table}

\begin{table}[!ht]
    \centering
    \scalebox{0.9}{
    \begin{tabular}{l|l}
    \toprule
        \textbf{Dataset} & \textbf{Evaluation Prompt} \\ \midrule
        TruthfulQA & \makecell[l]{\texttt{Judge the correctness of a given answer. Question: \{question\}\textbackslash n}\\\texttt{Answer: \{answer\}\textbackslash n Is the answer correct? Return Yes or No.}}\\ \midrule
        Alpaca &  \makecell[l]{\texttt{Please give helpful, very detailed, and polite answer to the user's question}\\ \texttt{below.\textbackslash n Question: \{question\}}} \\
    \bottomrule
    \end{tabular}}
    \caption{Dataset-specific prompts used for evaluation.}
    \label{tab:template2}
\end{table}

\section{Efficient Inference}
\label{appendix:vllms}

We implement the inference of our fused model on the existing inference framework, vLLM~\cite{kwon2023efficient}. Unlike other MoE models supported by vLLM, such as Mixtral~\cite{jiang2024mixtral}, our fused model requires different input prompts and the maintenance of multiple key-value caches within multiple models. Modifying the model implementation within vLLM directly to accommodate these requirements can be complex and may conflict with the PageAttention mechanism~\cite{kwon2023efficient} due to the use of multiple key-value caches. Therefore, we instead partition the GPU memory into several parts, each running a single model using a vLLM instance, and then fusing the output to form a fused model.

vLLM inherently supports streaming output, which returns tokens to the user-end token-by-token, and each token is produced by a sampler function applied on the hidden logits of the LLM. 
We change the implementation: in each iteration, we return the hidden logit instead of the token:
\begin{python}
# In model implementation
# change from outputting token = self.model.sample(hidden_states, sampling_metadata) to
return {
    "sampler": self.model.sample,
    "data": {
        "hidden_states": hidden_states,
        "sampling_metadata": sampling_metadata,
    }
}
\end{python}
This allows us to pause the model generation, giving us control over when to predict the next token and when to continue generating future tokens. We then make the model instances communicate and fuse the logits:
\begin{python}
logits = [
    llm.llm_engine.step()
    for llm in llms
] # get logits for different LLMs
fused_logits = fuse_function(*[logit["data"] for logit in logits]) # apply fuse function
\end{python}
The next token is predicted and sampled using the fused output, and we control the model instances to resume generation.


\section{Disscussion and Limitations}
\label{appendix:limitations}

We regard the data distribution in training language models into three domains in this study according to the symbol systems and achieve promising empirical results in our experiments. However, the realistic situation is far more sophisticated. In the field of ``text domain'' alone, there are different tasks such as common sense knowledge, specialized knowledge, natural language reasoning, etc., not to mention the existence of multilingualism. 
Our fused model may yield less favorable results on other benchmarks. 
In our training, no explicit selection mechanism is introduced in order to make the method scalable (force specialist models to process certain types of data). 
We believe finer-grained models could be trained under the spirit of \modelname; that is, the number of specialists is not necessarily three, and the domains are also necessarily divided as the same as the paper.
For example, other symbol systems (like DNA sequences) may also be integrated into the framework.

The \textsc{UltraChat} 2 dataset is fully synthetically generated by GPT-4 and fully excludes human engagement. Besides efficiency and privacy benefits, the factuality and trustworthiness of generated content can not be guaranteed. 
Our study mainly focuses on text, code, and math capabilities, neglecting some other important properties of LLMs, such as safety and multilinguality. We believe our proposed approach is generally applicable to address these limitations and will devote to developing more advanced methods.

Comparing to the line of works on model merging that manipulates the inner parameters of existing models in either supervised or unsupervised manner~\cite{daheim2023model, stoica2024zipit, wan2024knowledge,bansal2024llm}, our framework tackles the problem in a more straightforward way by directly merging the output and training with mixed high-quality instructional dataset to further adapt the model. The proposed framework follows the spirit of instruction tuning, and the training is conducted with direct supervised fine-tuning. Employing a diverse set of instruction datas, we show that the resulting model is equipped with desirable expertise and generalizes well to different domains of data. Moreover, our framework does not strictly require a similar model structure across specialists, and the structure design of the gating module on top of specialists can also be flexibly adjusted to match the desired learning capacity.


\section{Case Study }
\begin{figure*}[!ht]    
  \centering            
  \subfloat[Case study: tokens and weights of code data (a).]   
  {
      \label{fig:subfig1}\includegraphics[width=0.99\textwidth]{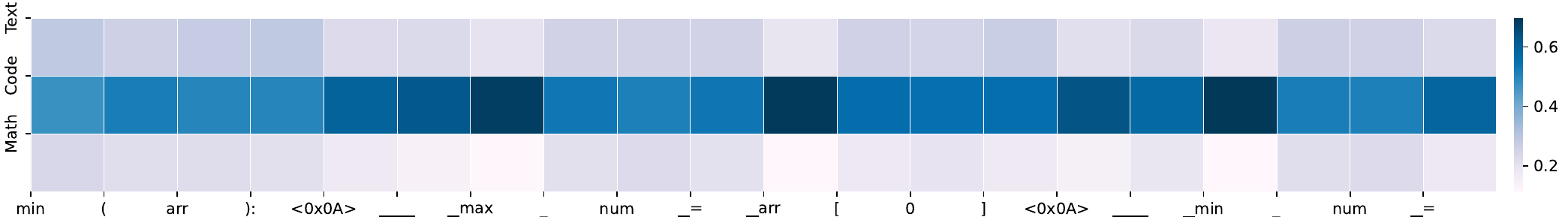}

  }
  \\
  \subfloat[Case study: tokens and weights of code data (b).]
  {
      \label{fig:subfig2}\includegraphics[width=0.99\textwidth]{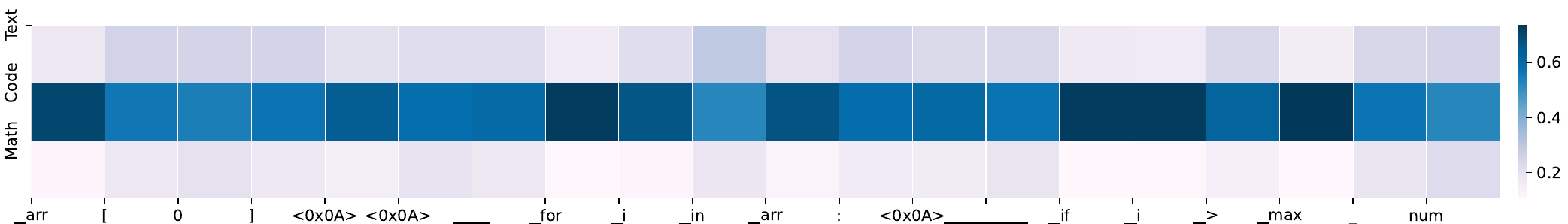}
  }
  \\
    \subfloat[Case study: tokens and weights of code data (c).]
  {
      \label{fig:subfig3}\includegraphics[width=0.99\textwidth]{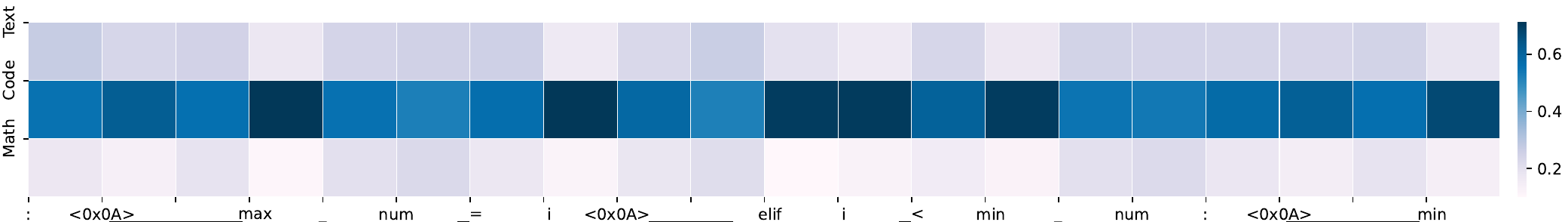}
  }
  \\
      \subfloat[Case study: tokens and weights of code data (d).]
  {
      \label{fig:subfig3}\includegraphics[width=0.99\textwidth]{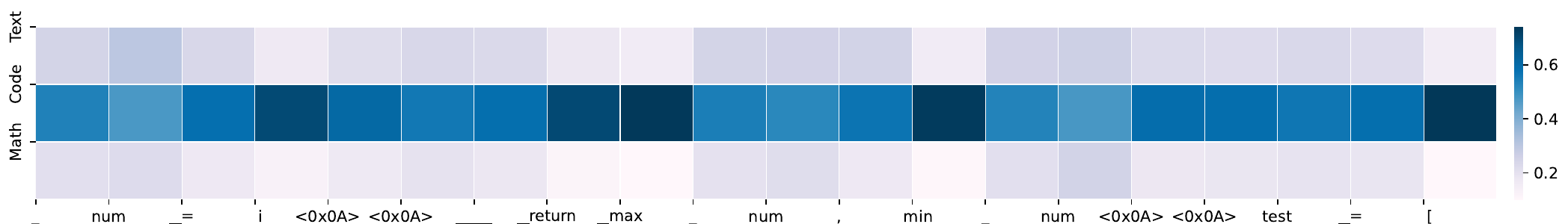}
  }
    \\
      \subfloat[Case study: tokens and weights of code data (e).]
  {
      \label{fig:subfig3}\includegraphics[width=0.99\textwidth]{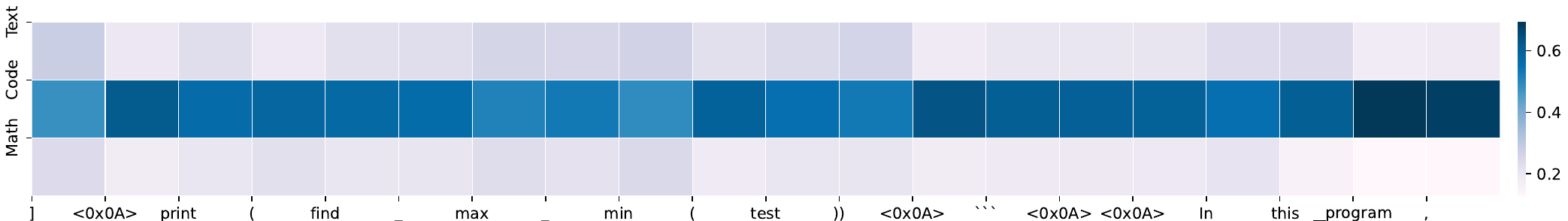}
  }
  \caption{Weight distributions of some pieces of tokens from a sample of code data.}  
  \vspace{0cm}
  \label{fig:case-code}            
\end{figure*}

\begin{figure*}[!th]    
  \centering            
  \subfloat[Case study: tokens and weights of math data (a).]   
  {
      \label{fig:subfig1}\includegraphics[width=0.99\textwidth]{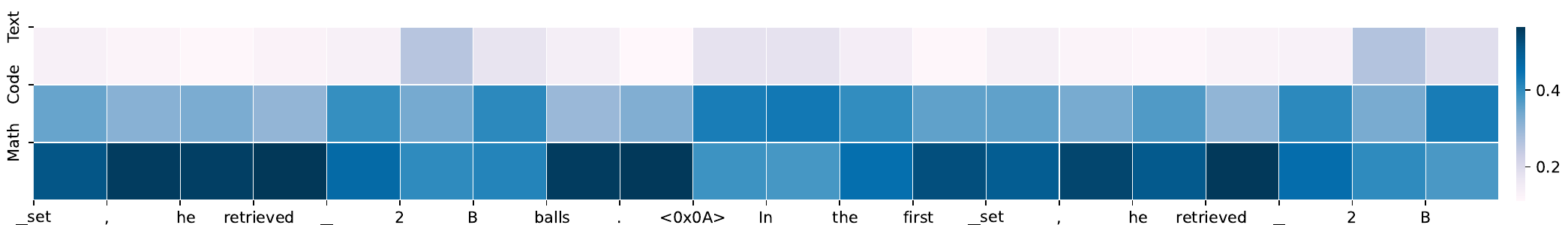}

  }
  \\
  \subfloat[Case study: tokens and weights of math data (b).]
  {
      \label{fig:subfig2}\includegraphics[width=0.99\textwidth]{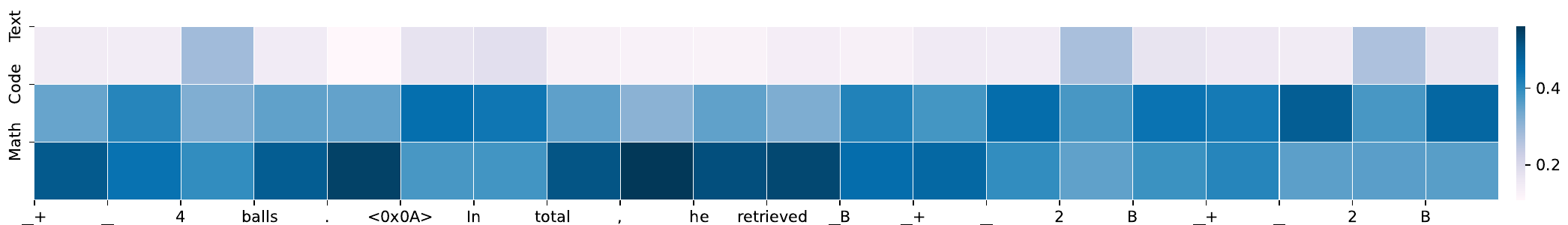}
  }
  \\
    \subfloat[Case study: tokens and weights of math data (c).]
  {
      \label{fig:subfig3}\includegraphics[width=0.99\textwidth]{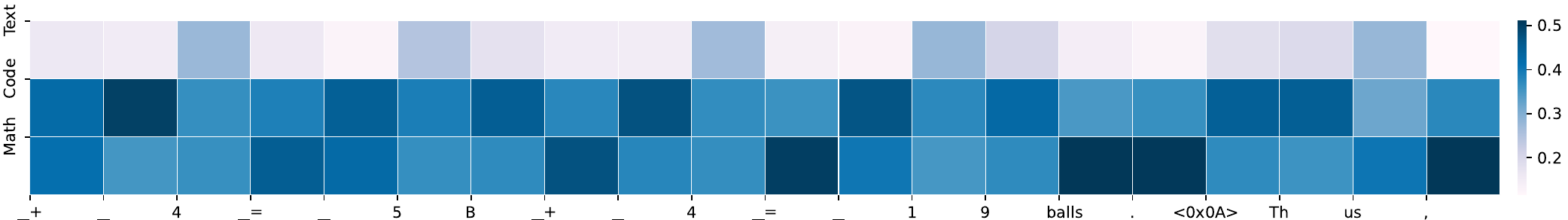}
  }
  \\
      \subfloat[Case study: tokens and weights of math data (d).]
  {
      \label{fig:subfig3}\includegraphics[width=0.99\textwidth]{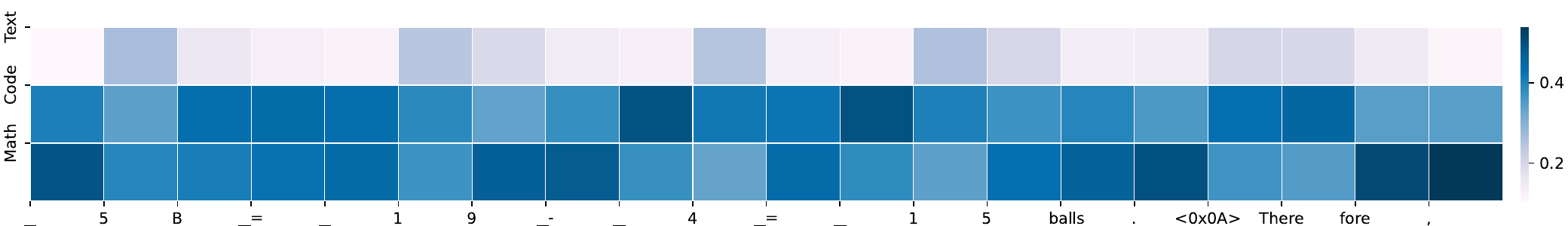}
  }
    \\
      \subfloat[Case study: tokens and weights of math data (e).]
  {
      \label{fig:subfig3}\includegraphics[width=0.99\textwidth]{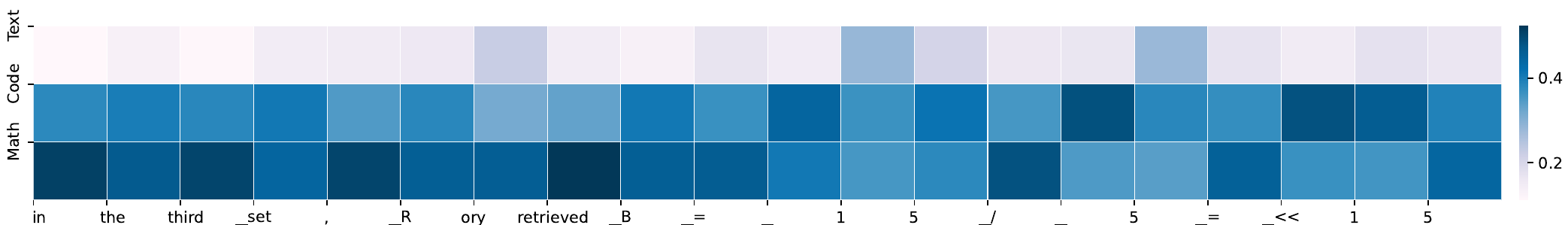}
  }
  \caption{Weight distributions of some pieces of tokens from a sample of math data.}  
  \vspace{0cm}
  \label{fig:case-math}            
\end{figure*}

In Section~\ref{sec:analysis}, we analyze the model expertise at the sequence level and set level. In this section, we provide cases at the token level to illustrate the weight distributions of the three specialist models.
Figure~\ref{fig:case-code} and Figure~\ref{fig:case-math} show two cases randomly extracted from \dataname\ code data and GSM8K dataset. For coding data, almost all weights are assigned to code specialist model. For math data, there is considerable weight given to code model as well, given the fact that mathematical equation is much alike code snippets. The assumption can be validated by the fact that when it comes to non-mathematical notation, the token weight distribution clearly favors the math specialist more.
The observation is in line with our expectation, that the fused model can implicitly learn to allocate tokens to suitable specialist to achieve better performance. Meanwhile, similarity between domains could be captured and their performance can be enhanced jointly by related specialists.

\end{document}